\documentclass{article}
\usepackage{spconf,amsmath,graphicx,hyperref}

\usepackage{xcolor}

\usepackage{bibentry}
\newcommand{\our}{\textcolor{black}{}{AutoContra}}

\title{Automated Prompt Generation for Creative and Counterfactual Text-to-Image Synthesis}
\name{Aleksa Jelaca$^{\dagger}$, Ying Jiao$^{\dagger}$$^{*}$\thanks{* Ying Jiao and Chang Tian are co-first authors.}, Chang Tian$^{\dagger}$$^{*}$\thanks{Corresponding authors: Chang Tian (namechangtian@163.com) and Ying Jiao (ying.jiao@kuleuven.be).}, Marie-Francine Moens$^{\dagger}$ }
\address{$^{\dagger}$ KU Leuven, Leuven, Belgium \\
\texttt{\{alekjelacamg000\}@gmail.com}}

\begin{document}
%
\maketitle
\begin{abstract}
Text-to-image generation has advanced rapidly with large-scale multimodal training, yet fine-grained controllability remains a critical challenge. Counterfactual controllability, defined as the capacity to deliberately generate images that contradict common-sense patterns, remains a major challenge but plays a crucial role in enabling creativity and exploratory applications. In this work, we address this gap with a focus on counterfactual size (e.g., generating a tiny walrus beside a giant button) and propose an automatic prompt engineering framework that adapts base prompts into revised prompts for counterfactual images. The framework comprises three components: an image evaluator that guides dataset construction by identifying successful image generations, a supervised prompt rewriter that produces revised prompts, and a DPO-trained ranker that selects the optimal revised prompt. We construct the first counterfactual size text–image dataset and enhance the image evaluator by extending Grounded SAM with refinements, achieving a 114\% improvement over its backbone. Experiments demonstrate that our method outperforms state-of-the-art baselines and ChatGPT-4o, establishing a foundation for future research on counterfactual controllability.
\end{abstract}
\section{Introduction}
\label{sec:intro}
Deep learning has significantly advanced numerous applications~\cite{antiover,fightingagainst,2024generic,2025using,tian2025large}, with text-to-image generation~\cite{paint4poem} representing an important area of progress.
Text-to-image generation is the task of synthesizing images from natural language descriptions, enabled by advances in large-scale multimodal training and diffusion models. Recent works (\cite{saharia2022photorealistic}, \cite{gani2023llm}, \cite{montenegro2024integrative}, \cite{OpenAI_Dalle3_2023}) have demonstrated remarkable success in producing photorealistic and stylistically diverse images that align with complex prompts, supporting applications in creative industries, design, education, and accessibility.

Despite progress, challenges persist in achieving fine-grained controllability over generated content. Existing research on controllable text-to-image generation has primarily focused on constraining specific visual aspects such as spatial layout \cite{zhang2023adding}, object attributes\cite{jiang2024comat}, style transfer \cite{sohn2023styledrop}, or local image editing \cite{yang2024mastering}. These methods are designed to enhance faithfulness to user prompts and improve alignment with real-world semantics. However, counterfactual controllability, which guides models to deliberately generate images that contradict common-sense statistical patterns, is underexplored. This gap is particularly relevant for creative, artistic, and exploratory applications, where generating counterfactual images serves not only as a test of model flexibility but also as a tool for fostering imagination beyond realistic distributions.

\begin{figure}
    \centering
    \includegraphics[width=\linewidth]{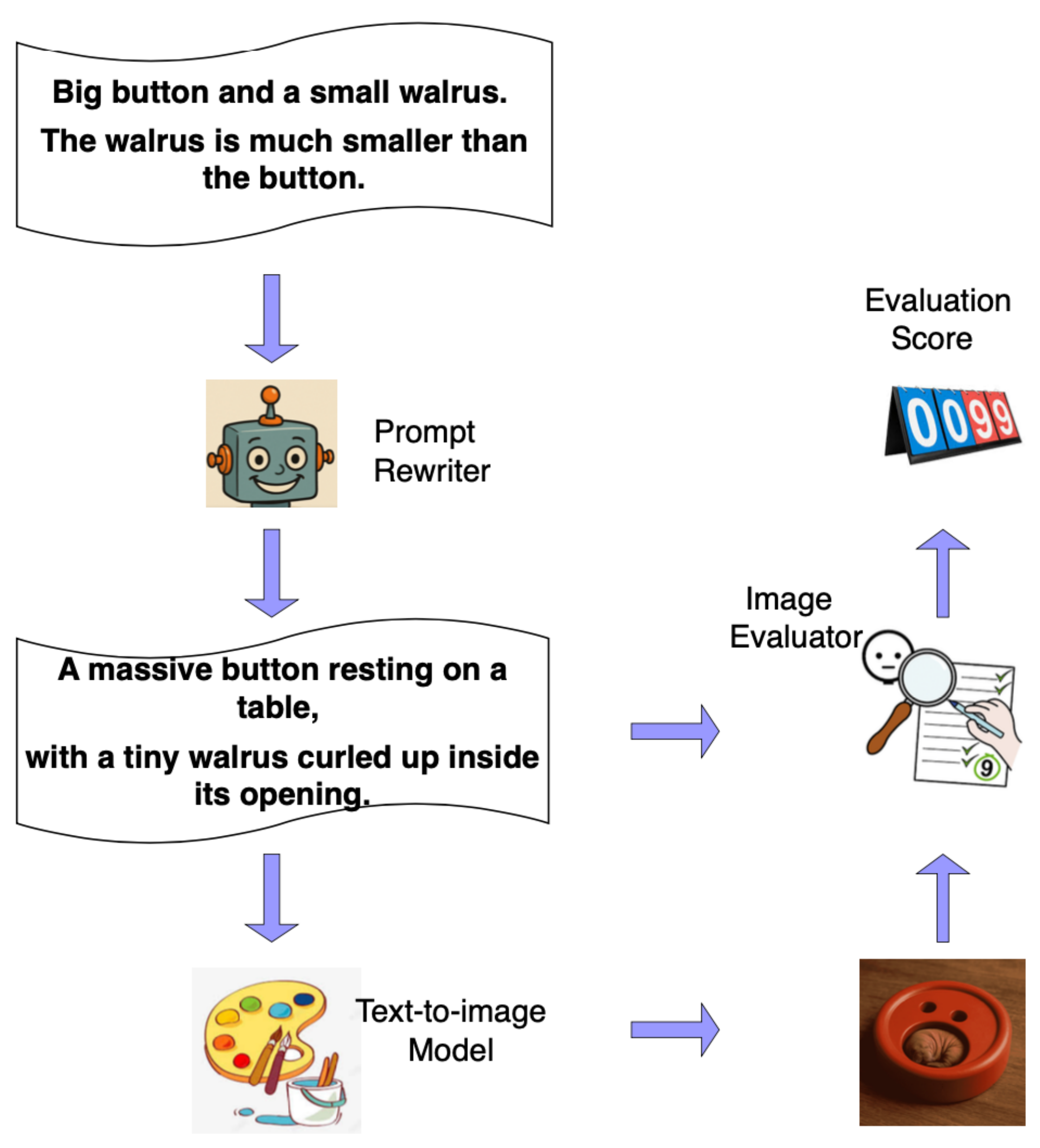}
    \caption{Overview of the proposed method.}
    \label{fig:pipeline}
\end{figure}

In this paper, we propose an automatic prompt engineering framework that adapts base prompts to revised prompts that lead to images faithful to counterfactual requirements as show in Figure \ref{fig:pipeline}. 
This work investigates counterfactual size as a primary research focus (e.g., generating an image of a giant button next to a tiny walrus), while the proposed method is readily extensible to other counterfactual attributes.
Our framework contains three components: an image evaluator, a prompt rewriter, and a prompt ranker. Since no dataset of prompts paired with counterfactual size images exists, we construct one with the help of the image evaluator. The image evaluator measures the degree to which images satisfy counterfactual size requirements, enabling us to classify prompts into successful ones (which yield faithful counterfactual images) and failed ones. Our prompt rewriter is a pretrained language model fine-tuned with supervised learning on the successful prompt set. The prompt ranker is another pretrained language model fine-tuned with Direct Preference Optimization (DPO) \cite{rafailov2023direct} on both successful and failed prompts. At inference time, the fine-tuned prompt rewriter generates multiple candidate prompts, and the reranker selects the top candidate as the final revised prompt.

We conduct experiments with the open-source text-to-image model CoMat \cite{jiang2024comat}. The results demonstrate that our framework outperforms state-of-the-art automatic prompt rewriting baselines for text-to-image generation, and ChatGPT-4o \cite{OpenAI_ChatGPT-4o_2024} when used as a prompt rewriter.

Our contributions are as follows~\footnote{The code and data will be publicly available upon publication at https://github.com/jekia2000/Counterfactual-Size-T2I.}:
\begin{itemize}
    \item We present an in-depth investigation of counterfactual controllability in text-to-image generation, with a primary focus on counterfactual size.
    \item We design an automatic prompt engineering framework that combines an evaluator-guided dataset construction process, a supervised fine-tuned rewriter, and a DPO-trained ranker.  
    \item We build the first dataset of counterfactual size prompts and images, facilitating future study of this new task.  
    \item We experimentally show that our framework improves counterfactual text-to-image generation with respect to size compared to existing baselines.
\end{itemize}

\section{Proposed Method}
\label{sec:methodolody}

\subsection{Image Evaluator}
\label{subsec:evaluator}

\begin{figure}[h]
    \centering
    \includegraphics[width=1.0\linewidth]{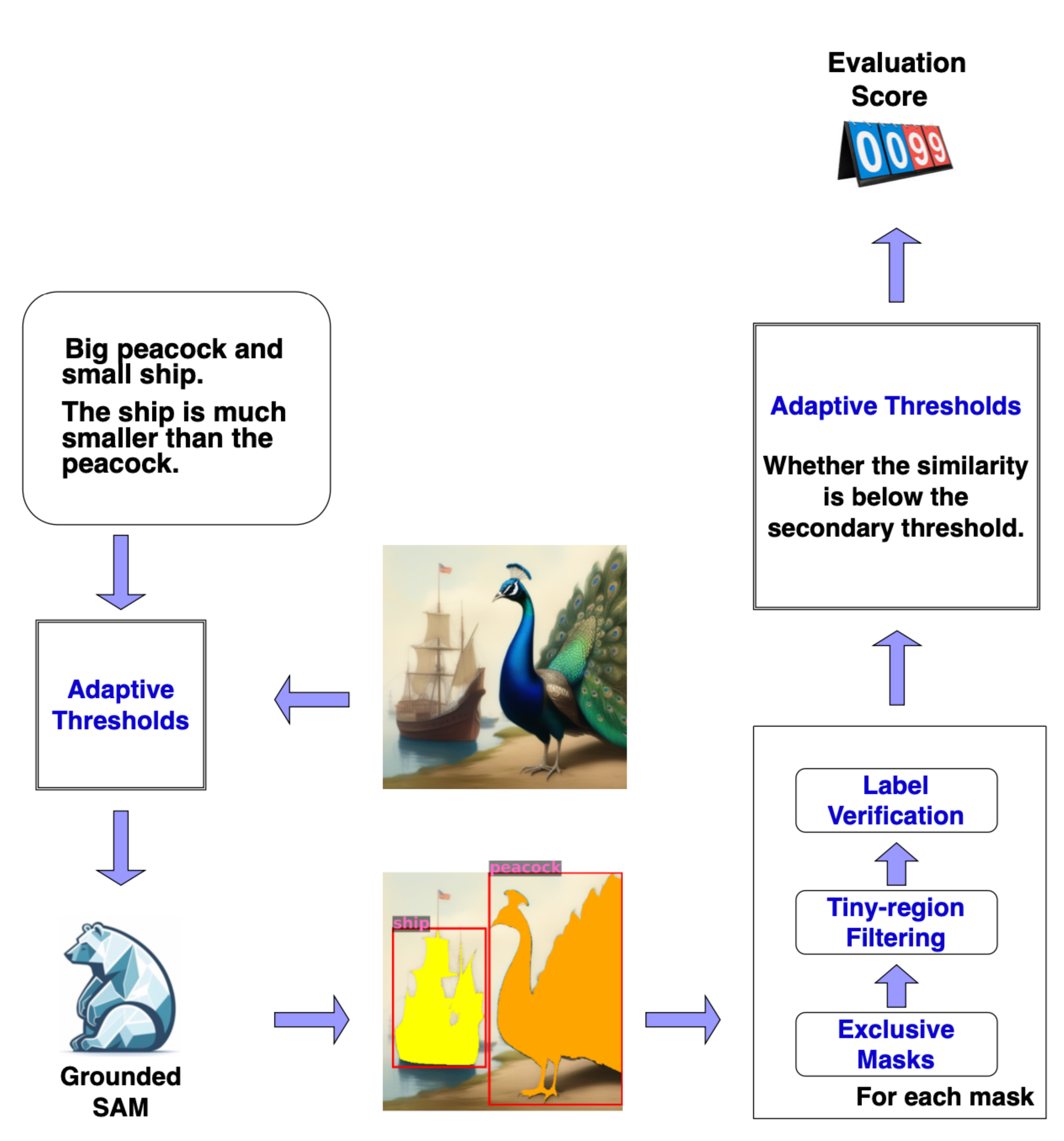}
    \caption{The holistic view of the Image Evaluator Components. The detected mask areas for the objects are presented in the figure.}
    \label{fig:evaluator}
\end{figure}

We design an evaluator that assigns a score reflecting how well an image generated from a text prompt depicts a counterfactual size relationship between two objects: one that is typically large in reality (the \emph{large} object) and one that is typically small (the \emph{small} object). A higher score indicates better alignment with the counterfactual requirement that the small object appears larger than the big one. An overview of the evaluator is shown in Figure~\ref{fig:evaluator}.

The evaluator relies on segmentation masks from Grounded SAM \cite{ren2024grounded}. Each mask is associated with a text label and has a measurable pixel area. To align the segmentation with human judgment, four refinements are introduced:

\textbf{Tiny-region filtering.} Detections with very few pixels are discarded as noise.

\textbf{Exclusive masks.} Grounded SAM may output overlapping masks, where one covers both the small and big objects. To prevent this, masks are sorted by area (smallest first), and pixels assigned to smaller masks are excluded from larger ones. This guarantees mutual exclusivity and avoids a single mask absorbing both objects.
    
\textbf{Label verification.} Grounded SAM may mislabel objects. To correct this, the image region corresponding to each mask is extracted, placed on a blank background, and then embedded with CLIP \cite{radford2021learning}. The embedding is compared against a reference database of segmented web objects, and the label with maximum cosine similarity is assigned:
\begin{equation}
\label{eq:vector-db}
\text{label}^{\star} = \arg\max_{(l,e)\in \mathcal{D}} \cos\!\left(f(O_m), e\right),
\end{equation}
where $O_m$ is the image region from mask $m$, $f(\cdot)$ the CLIP encoder, and $\mathcal{D}$ the reference label–embedding database. $l$ denotes a candidate label in the reference database, and $e$ represents the embedding corresponding to label $l$.
     
 \textbf{Adaptive thresholds.} Grounded SAM predictions rely on both box and text thresholds. Higher thresholds reduce noise but may miss true objects, whereas lower thresholds detect more regions at the risk of introducing irrelevant predictions.
  To balance this, thresholds $(\tau_{\text{box}}, \tau_{\text{text}})$ are dynamically adjusted based on the CLIP similarity between the generated image $I$ and the textual description $T$. The CLIP similarity here refers to the cosine similarity between the embeddings of the image and the text, computed using the CLIP model:

\begin{equation}
(\tau_{\text{box}}, \tau_{\text{text}}) = 
\begin{cases} 
(b_l, t_l), & \cos\_sim(I, T) \ge \mu_a \\ 
(b_g, t_g), & \text{otherwise}
\end{cases}
\end{equation}

where $b_l, t_l$ and $b_g, t_g$ denote lower and greater box/text thresholds, respectively, and $\mu_a$ is the main similarity cutoff. Additionally, a secondary similarity threshold $\mu_b$ handles extreme cases: if $\cos\_sim(I, T)$ falls below this threshold, the evaluator assumes only one object is present. 

After refinement, the evaluator assigns a score based on object presence and relative size:
\begin{equation}
\label{eq:score}
S = 
\begin{cases}
\min\!\left(\tau_R, \frac{A_s}{A_b}\right), &
\text{both present, size ratio correct}, \\[1mm]
\max\!\left(-\tau_R, -\frac{A_b}{A_s}\right), &
\text{both present, size ratio incorrect}, \\[1mm]
-\tau_R*(1+g), &
\text{one object missing}, \\[1mm]
-\tau_R*(1+g)^2, &
\text{both objects missing},
\end{cases}
\end{equation}
where $A_s$ and $A_b$ denote the largest detected mask areas of the small and big objects, respectively, in the real scene. $\tau_R$ is the clipping threshold, and $g$ is a penalty factor. Positive scores are clipped above $\tau_R$, reflecting that once counterfactuality is established, further increases in the size ratio do not matter. Negative scores are similarly clipped.

\subsection{Dataset Construction}
\label{subsec:dataset}

Since no public dataset exists for counterfactual size prompt–image pairs, we construct a dataset of 91 objects: 46 typically large and 45 typically small. Large objects include animals, vehicles, and monuments, while small objects comprise animals, household items, clothing, accessories, and footwear. Pairing each large with each small object yields 2070 base prompts using the template: 
``Big [small object] and small [big object]. The [big object] is much smaller than the [small object].'' 

We use twelve manually written prompts producing faithful counterfactual-sized images as few-shot examples for ChatGPT-4o to generate revised prompts. Each revised prompt is used to generate an image with CoMat SDXL~\cite{jiang2024comat} and scored by the Image Evaluator. Prompts exceeding a reward threshold ($\tau_{\text{reward}}$) are labeled positive, while those with negative scores are labeled negative. From these, we build a dataset of 7304 triplets—each consisting of a base prompt, a positive, and a negative rewritten output—for DPO finetuning of the prompt ranker. For supervised finetuning of the prompt rewriter, we extract base–positive pairs.

\subsection{Training and Inference of Prompt Rewriter and Ranker}

The prompt rewriter is fine-tuned via supervised learning on the base–positive prompt pairs constructed in \ref{subsec:dataset}, learning to transform base prompts into revised candidates likely to generate faithful counterfactual size images. The prompt ranker is separately fine-tuned using DPO on triplets consisting of a base prompt, a positive, and a negative rewritten output, learning to assign higher scores to more effective rewrites.

During inference, the fine-tuned prompt rewriter produces multiple candidate revisions for each base prompt. The prompt ranker then evaluates these candidates, and the one with the highest score is selected as the final output. This two-stage framework combines the generative capacity of the rewriter with the discriminative power of the ranker, ensuring that the selected prompt consistently produces images that faithfully reflect the intended counterfactual size relationships.

\section{Experiments}
\label{sec:experiment}
\subsection{Implementation details}

For the image evaluator, we set the height/width threshold for tiny objects to 32 pixels and the mask area threshold to 2048. Other parameters include $b_l=0.2$, $t_l=0.2$, $b_g=0.3$, $t_g=0.25$, $\mu_a=0.39$, $\mu_b=0.33$, $\tau_{R}=1.5$, and $g=0.5$. The backbone model of the prompt rewriter is GPT2 \cite{radford2019language}. We finetune it with a learning rate of $5\times10^{-5}$, a batch size of 16, maximum sequence length of 64, for up to 20 epochs with patience of 3 epochs. Optimization uses AdamW \cite{loshchilov2017decoupled} with gradient accumulation steps of 1. The backbone model of the prompt ranker is GPT2-large. Finetuning uses a learning rate of $5\times10^{-6}$, a batch size of 64, maximum sequence length of 64, and runs for 20 epochs. We set $\beta=0.1$, ref\_model\_mixup\_alpha=1, and ref\_model\_sync\_steps=512. We use AdamW as the optimizer for all finetuning. During inference, the rewriter generates 15 candidate prompts per base prompt by non-deterministic sampling with temperature of 0.6 and nucleus sampling of 1. 

All experiments were conducted with three random seeds (40, 41, and 42), each requiring at least 32 GiB of GPU memory \footnote{\url{https://www.hpc.cineca.it/}}\footnote{\url{https://www.vscentrum.be/}}. Further details are available in our GitHub repository.

\subsection{Research Questions}

Our experiments are designed to address the following research questions:

\textbf{RQ1}: Can our prompt rewriting framework effectively lead to faithful counterfactual size images, and how does its performance compare with baseline methods?

\textbf{RQ2}: Does incorporating the prompt ranker improve the quality of rewritten prompts and the resulting images, compared to using the prompt rewriter alone?

\textbf{RQ3}: To what extent does the automatic image evaluator align with human judgments of counterfactual size faithfulness?

\subsection{RQ1: Effectiveness of \our}
To answer RQ1, we compare \our{} against two baselines: Promptist \cite{hao2023optimizing}, an automated prompt optimization method for text-to-image generation, and ChatGPT-4o, used directly as a prompt rewriter. We additionally report results for the template base prompts. Accuracy is measured as the proportion of images scoring at least $\tau_{\text{reward}}$ in the image evaluator. We use a test set of 1004 object pairs.

As shown in Table~\ref{tab:evaluation}, Promptist performs the worst, likely because its strategy of adding stylistic descriptors obscures the intended object–size relationships, preventing the model from capturing counterfactuality. In contrast, ChatGPT-4o achieves much stronger results, demonstrating the capability of large language models to rewrite prompts in ways that align with non-trivial semantic constraints. Our proposed AutoContra framework outperforms both baselines. Nevertheless, the overall accuracy remains relatively low, underscoring the inherent difficulty of this task and the significant room for improvement in counterfactual controllability.

Figure \ref{fig:examples} presents example images generated from different prompt rewriting strategies along with their image evaluator scores. These qualitative results complement the quantitative analysis, showing how \our{} better capture the intended counterfactual relationships.

\begin{table}[h] 
\centering

\begin{tabular}{|c|c|}
  \hline
   \textbf{Method} & \textbf{Accuracy (\%)} \\ \hline
   Base Prompts & $10.2 \pm 0.9$ \\ \hline
 Promptist & $8.2 \pm 0.3$  \\ \hline
 ChatGPT-4o & $27.5 \pm 1.7$ \\ \hline
 AutoContra w.o. Ranker & $29.1 \pm 0.7$  \\ \hline
 AutoContra (ours) & $\textbf{30.3} \pm \textbf{0.8}$ \\ \hline
 
\end{tabular}
  \caption{Evaluation results with three seeds.}
\label{tab:evaluation}
\end{table}

\subsection{RQ2: Contribution of the Prompt Ranker}
\begin{figure}
    \centering
    \includegraphics[width=1\linewidth]{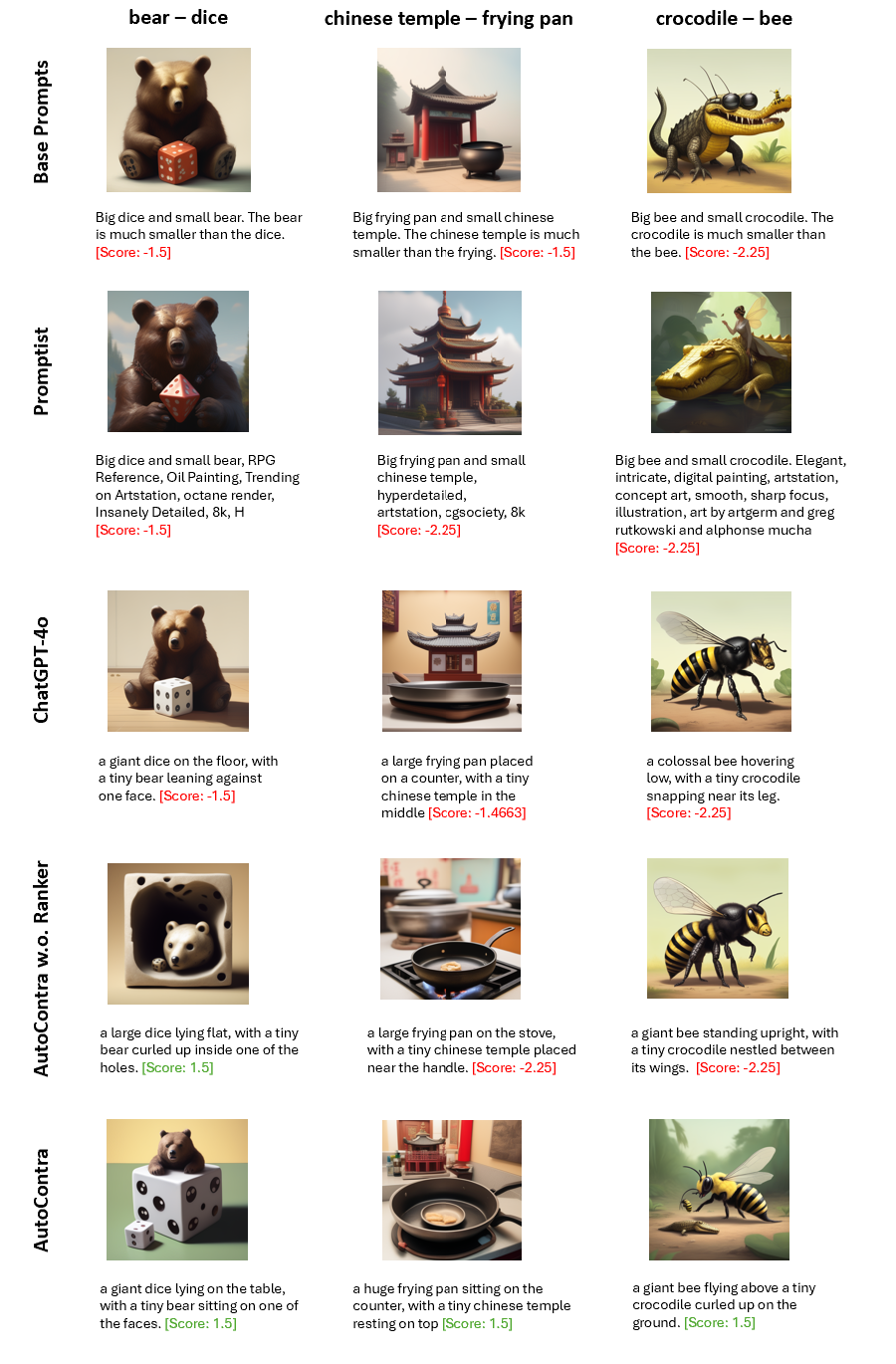}
    \caption{Visual Comparison of images of different object pairs across different prompt rewriting techniques}
    \label{fig:examples}
\end{figure}

As shown in Table~\ref{tab:evaluation}, adding the prompt ranker improves accuracy from $29.1\%$ to $30.3\%$, demonstrating its positive effect on performance. Qualitative examples in Figure \ref{fig:examples} further support this finding: without the prompt ranker, some generated images contain only one object, whereas with the prompt ranker, both objects are more consistently present and correctly follow the counterfactual size requirement.

\subsection{RQ3: Ablation Study of Image Evaluator}
\label{subsec:ablationstudy}
Our image evaluator is built on Grounded SAM with several refinements, as described in Section~\ref{subsec:evaluator}. To assess the impact of each component, we conduct an ablation study, with results summarized in Table~\ref{tab:ablation}.  

Experiments are performed on a dataset of 235 diverse images generated from 50 distinct small--large object pairs, each representing a class category defined in Section~\ref{subsec:dataset}. Human annotators labeled each generated image as \textit{counterfactual size} (True/False), and these annotations serve as ground truth.  

We then apply different variants of the image evaluator to predict whether a generated image exhibits counterfactual size and compare the predictions against human annotations. The main objective is to evaluate whether the evaluator correctly identifies all images that would receive a positive reward according to human judgment. Importantly, we focus on detecting positive rewards rather than enforcing the maximum threshold ($\tau_{R}=1.5$), since humans typically assess size relations visually rather than estimating precise ratios. Performance is measured using the F1 score.  

The results highlight several insights:
    (1) Removing adaptive thresholds (W.o. Adaptive thresholds) reduces evaluator performance, as Grounded SAM either masks irrelevant candidates as objects or misses true objects without appropriate threshold constraints.  
    (2) Excluding both adaptive thresholds and label verification (W.o. Adaptive thresholds + Label verification)  significantly degrades performance, demonstrating the necessity of label verification. Grounded SAM is effective at proposing candidate regions but less accurate at classification, whereas our reference embedding database provides specialized external knowledge to support accurate classification.  
    (3) Removing all refinements (W.o. All refinements) leads to substantial performance deterioration, with the plain evaluator achieving only 0.41 F1 compared to 0.88 for the complete evaluator. Overall, these findings confirm the reliability of our image evaluator and its strong alignment with human judgments.

\begin{table}[h] 
\centering
\begin{tabular}{|c|c|}
  \hline
  \textbf{Model} & \textbf{F1 score} \\ \hline
  Image Evaluator (ours)    & \textbf{0.882353}   \\ \hline
  W.o. Adaptive thresholds      & 0.809756   \\ \hline
  W.o. Adaptive thresholds+Label verification      & 0.441717   \\ \hline
 W.o. All refinements    & 0.411428   \\ \hline
  \end{tabular}
  \caption{F1 scores across model variants. W.o. means without. W.o. All refinements indicates that only Grounded SAM is used as the image evaluator variant.
}
\label{tab:ablation}
\end{table}
\section{Conclusion}
\label{sec:conclusion}
This work investigates counterfactual controllability in text-to-image generation, with counterfactual size as the primary focus. We introduce an automatic prompt engineering framework that integrates an image evaluator-guided dataset construction process, a supervised prompt rewriter, and a DPO-trained prompt ranker. By extending Grounded SAM with adaptive refinements, the image evaluator demonstrates substantial improvements in identifying faithful counterfactual generations. Leveraging this evaluator, we construct the first counterfactual size dataset, enabling systematic study of this underexplored task. Experimental results showed that our framework consistently outperforms state-of-the-art baselines and ChatGPT-4o, achieving significant gains in performance and alignment with human judgments. We believe this work establishes a foundation for future research on counterfactual image synthesis, contributing both methodological advances and resources that support creative, artistic, and exploratory applications.

\vfill\pagebreak

\bibliographystyle{IEEEbib}
\bibliography{refs}

\end{document}